\documentclass[conference]{IEEEtran}
\IEEEoverridecommandlockouts

\usepackage{cite}
\usepackage{amsmath,amssymb,amsfonts}
\usepackage{algorithmic}
\usepackage{graphicx}
\usepackage{textcomp}
\usepackage{xcolor}
\def\BibTeX{{\rm B\kern-.05em{\sc i\kern-.025em b}\kern-.08em
    T\kern-.1667em\lower.7ex\hbox{E}\kern-.125emX}}

\ifCLASSINFOpdf
\else
\fi

\usepackage{cite}
\usepackage{amsmath,amssymb,amsfonts}
\usepackage{algorithmic}
\usepackage{graphicx}
\usepackage{textcomp}
\usepackage{xcolor}
\usepackage{rotating}
\def\b{\ensuremath\boldsymbol}

\usepackage{url}
\newcommand{\R}{\mathbb{R}}
\usepackage[algo2e,linesnumbered,lined,boxed,vlined]{algorithm2e}
\usepackage{multirow}

\begin{document}

\title{ Hierarchical Subspace Learning for Dimensionality Reduction to Improve
Classification Accuracy in Large Data Sets }

\author{Parisa Abdolrahim Poorheravi,Vincent Gaudet
       
% \hfill\break

Department of Electrical and Computer Engineering, University of Waterloo, Waterloo, ON, Canada
        
Emails: \{pabdolra, vcgaudet\}@uwaterloo.ca

}

\author{\IEEEauthorblockN{Parisa Abdolrahim Poorheravi}
\IEEEauthorblockA{\textit{Department of Electrical and Computer Engineering} \\
\textit{University of Waterloo}\\
Waterloo, ON, Canada \\
pabdolra@uwaterloo.ca}
\and
\IEEEauthorblockN{Vincent Gaudet}
\IEEEauthorblockA{\textit{Department of Electrical and Computer Engineering} \\
\textit{University of Waterloo}\\
Waterloo, ON, Canada \\
vcgaudet@uwaterloo.ca}
}

\maketitle

\begin{abstract}

Manifold learning is used for dimensionality reduction, with the goal of finding a projection subspace to increase and decrease the inter- and intraclass variances, respectively. However, a bottleneck for subspace learning methods often arises from the high dimensionality of datasets. In this paper, a hierarchical approach is proposed to scale subspace learning methods, with the goal of improving classification in large datasets by a range of 3\% to 10\%. Different combinations of methods are studied. We assess the proposed method on five publicly available large datasets, for different eigen-value based subspace learning methods such as linear discriminant analysis, principal component analysis, generalized discriminant analysis, and reconstruction independent component analysis. To further examine the effect of the proposed method on various classification methods, we fed the generated result to linear discriminant analysis, quadratic linear analysis, k-nearest neighbor, and random forest classifiers. The resulting classification accuracies are compared to show the effectiveness of the hierarchical approach, reporting results of an average of 5\% increase in classification accuracy.
\end{abstract}

% \begin{IEEEkeywords}

% \end{IEEEkeywords}

\IEEEpeerreviewmaketitle
%%%%%%%%%%%%%%%%%%%%%%%%%%%%%%%%%%%%%%%%%

\section{Introduction}
Data sets used in today's machine learning algorithms are becoming more and more complex, often incorporating a large number of features. However, having too many features in a data set can hinder intuition about a problem and over-complicate information processing. Indeed, having a high number of features compared to the number of samples in a dataset is sometimes called the ``curse of dimensionality". To help with visualization and exploration in such scenarios, dimensionality-reduction techniques can be used to pre-process large data sets, as illustrated in a prior ISMVL paper\cite{suda2018systematic}. 

Dimensionality reduction is a method that removes redundant and irrelevant features. There are two main approaches for reducing the size of a dataset: feature selection \cite{chandrashekar2014survey} and feature extraction\cite{sorzano2014survey}. Feature selection chooses a subset of features with respect to their correlation with labels. It does not change the feature space and it is most effective when intrinsic complexity of data is lower than the complexity of its feature space. Some of the well-known approaches are chi-square \cite{witten2002data}, correlation based \cite{witten2002data}, and sparse clustering based feature selection \cite{witten2010framework}. Feature extraction however, can reduce the dimensionality of data by changing the data space and creating an entirely new feature set that best represents the data, albeit in a lower-dimension data space. Some of the well-known approaches are principal component analysis
(PCA) \cite{jolliffe2016principal}, linear discriminant analysis (LDA) \cite{duda2006pattern}, and LDA/MMC (maximum margin criterion) \cite{liu2014scatter}.

Dimensionality reduction often produces a more interpretable representation of data. However, it does not necessarily result in higher accuracy due to potentially losing some informative features during the reduction. However, it offers a trade-off to gain better explorability and visualization of data. In this paper, inspired by a hierarchical approach for large-margin metric learning using stratified sampling \cite{poorheravi2020acceleration}, we propose a hierarchical approach for manifold learning dimensionality reduction. This approach includes iterative selection of features by hierarchical sampling, feature selection, and feature extraction. 

The remainder of this paper is organized as follows. In section \ref{section_literature_review}, we review feature selection and feature extraction methods used in the hierarchical approach. We discuss the classifiers that were used to compare the accuracies of the methods under study. The proposed hierarchical method is discussed in more detail in section \ref{proposed_method}. Section \ref{section_experiments} outlines the different combinations and experiments of the hierarchical approach. It also compares the results of this proposed approach with the original manifold learning method and raw data. Finally, section \ref{section_conclusion} concludes the paper and discusses possible future directions.

%%%%%%%%%%%%%%%%%%%%%%%

\section{Background}\label{section_literature_review}

\subsection{Feature Selection}
Feature selection \cite{chandrashekar2014survey,miao2016survey,cai2018feature} is a pre-processing dimensionality-reduction method to select a subset of features that represents the data while preserving its classification. The goal is to map a complex raw dataset denoted by $\b{X}\in \R^{n\times d}$ to a lower dimension of  $\b{X}\in \R^{z\times d}$ where $z \le n$, where $n$ and $z$ refer to numbers of features, and $d$ is the number of instances or cases in the data set. 

In this paper, correlation-based feature selection with respect to classes is used, which calculates the correlation of each feature based on its relevance to the label vector. A feature with a higher correlation value is more relevant to class determination, and therefore is ranked higher in the hierarchy of selected features \cite{ghojogh2019feature}.

\subsection{Feature Extraction and Dimensionality Reduction}
Feature extraction is a pre-processing dimensionality-reduction method that creates an entire set of new features based on, but different from, the original dataset \cite{sorzano2014survey}. Feature extraction considers the correlation between all features when mapping from $\b{X}\in \R^{n\times d}$ to a lower dimension of  $\b{Y}\in \R^{z\times d}$ where $z < n$ \cite{carreira1997review}. This mapping also changes the space of the dataset, allowing for a better representation or better visualization of the dataset, hence the term subspace learning methods. This newly created smaller set of features is called an embedded feature space. In this paper, a variety of supervised and unsupervised feature-extraction methods are used as described below.

\subsubsection{Principal Component Analysis (PCA)}
 is a linear unsupervised method for feature extraction that reduces the dimensionality of data by mapping it to a subspace with a lower dimension \cite{pearson1901liii , wold1987principal , abdi2010principal}. The goal of PCA is to find orthogonal directions in the space of data that best capture its variations. The maximum variation of data is called the first principal component and is denoted by $\b{u}$. PCA then projects the data such that the main coordinate is in the direction of $\b{u}$ by $\b{u^TX}$. The variance of the points after projection will then be $\b{u^TSu}$ where $\b{S}$ is the covariance matrix of data. PCA tries to maximise this variance, which leads to finding the leading eigenvectors of the covariance matrix called principal components denoted by $\b{U}$ with dimension of $1\times d$. The first $p$ terms of $\b{U}$ are chosen for dimensionality reduction.

\subsubsection{Independent Component Analysis (ICA)}
 is an extension of PCA that helps with non-Gaussian data.

\subsubsection{Fisher Discriminant Analysis (FDA)}
 also known as Linear Discriminant Analysis (LDA), is a supervised feature-extraction method that projects data on a new space with lower dimension based on an eigenvalue resolution\cite{duda2006pattern} \cite{fisher1936use, xu2006analysis, zhao1999subspace}. Similar to PCA, FDA uses the projection of data along the data direction, but it also incorporates the label information to find the optimum variation of data that increases and decreases the intra- and inter-class scatter, respectively \cite{ghojogh2019feature}.  If we denote the intra-class scatter as $S_b$, inter-class scatter as $S_w$ and the projection direction by $u$, we can formulate the FDA as \cite{brooks2005vibration}:
\begin{equation}
\begin{aligned}
J(u) := \frac{u^TS_bu}{u^TS_wu}
\end{aligned}
\end{equation}

FDA maximizes $J(u)$ to find the projection directions which are eigen vectors of $S_w^{-1}S_b$ sorted in descending order.

\subsubsection{Generalized Discriminant Analysis (GDA)}
 deals with non-linear discriminant analysis problems using kernel functions  \cite{baudat2000generalized ,haghighat2015cloudid}. The concept of GDA is the same as LDA, i.e., to maximize the ratio of between-class scatter to within-class scatter while mapping the data into a low-dimensional space.

\subsection{Classification Methods}
Classification is a task of assigning a class $y_j$ where $1<j<K$ to an input test data $x_i$ where $i \in \R^{n}  $ based on its attributes and the learning model that is achieved from the training set. The training data are used to create a model to calculate the probability of a point belonging to a class with respect to its features. In overall, the decision is based upon the calculated probability of a test sample belonging to a class. Variety of methods exist to create that classification model such as decision trees or neural networks. In this section we go through the methods we used for determining the accuracy of our proposed approach.  
\subsubsection{ k - Nearest Neighbor Classification (k-NN)}
 uses the known labels of the closest points in a training set to predict the class of each data point. In this paper, we use Euclidean distance metric which does not consider any weights for the points and values all the of them equally. 

If we denote a sample in testing set as $x_i$,  to predict its label $y_i$,  we determine the $k$ nearest neighbours of $x_i$ in the training set $T$. Then a majority vote of the known labels of $T$ gives us the predicted $y_i$.

\subsubsection{Linear Discriminant Analysis Classification (LDA)} calculates the probability of a point belonging to a class based on Bayes classification \cite{hastie2009elements,lachenbruch1979discriminant,tharwat2017linear}. LDA assumes that class conditional distributions are Gaussian and all classes share the same co-variance matrices. LDA calculates  the Euclidean distance of the testing set with respect to eigen vectors learned from the training set to predict the appropriate class. The closest eigen vector to the test point determines its predicted label. Assuming the class conditional distribution is Gaussian, then the mean and the covariance matrix of each class can be calculated. It results in a linear decision boundary to distinct classes. 

\subsubsection{Quadratic Discriminant Analysis Classification (QDA)}
is the quadratic form of LDA \cite{hastie2009elements ,lachenbruch1979discriminant}. Similar to LDA, it is based on Bayes classification with Gaussian class distributions, but unlike LDA, it does not assume the co-variance matrices of classes are the same. This assumption results in quadratic decision boundaries. 

\subsubsection{Random Forest Classification (RF)}
 is an ensemble learning method that creates a forest of decision trees, each independently created by randomly sampling a subset of data points in terms of both instances and features from training set \cite{liaw2002classification , criminisi2012decision}. Test data are passed through all trees, and the final decision is based on a majority vote among predictions.

\subsection{Hierarchical Large Margin Metric Learning with Stratified Sampling} \label{section_hierarchical}

A hierarchical approach with stratified sampling was proposed for accelerating large margin metric learning that suggested using portions of training data in a hierarchical manner to solve semi-definite programming optimization at every iteration \cite{poorheravi2020acceleration}. As the title suggests, this approach is iterative. The general idea was to consider points within some hypersphere in the data space and to sample from those points utilizing stratified sampling in an iterative manner. 

In stratified sampling, the data to be sampled is divided to homogeneous sub-groups called strata. Random sampling is applied on each strata \cite{barnett1974elements}.

Using the sample points from every hypersphere, semidefinite programming optimization is done and therefore the pace of optimization significantly improves. 

Initially, many small hyperspheres are considered on the data space to make small changes to the groups of data points. Triplets are sampled in hyperspheres using stratified sampling, considering classes as strata.Then, SDP optimization is solved over the sample points rather than the whole data. 

By solving the optimization in every hypersphere, a projection matrix is found to project all data points into the new metric subspace that is learned from sampled triplets. In later iterations, the number of hyperspheres is reduced, but their radius is increased in order to explore more of the data space. Moreover, at every iteration, the sampling portion is decreased because more data points fit into large hyperspheres, which can slow down the optimization.

\section{Proposed Hierarchical Subspace Learning Method} \label{proposed_method}

The main problem with non-linear dimensionality reduction in manifold learning is the high number of features found in complex datasets. Visualization, representation and exploration of high-dimension data is often non-intuitive to people interpreting the data. Hence, we can try to reduce the dimensionality of data while trying to preserve its structure. Inspired by the hierarchical large margin metric learning method with stratified sampling described above, we propose a hierarchical subspace learning method for better discrimination of classes.

The general idea of this approach is to iteratively train the algorithm with portions of data to improve performance. The procedure is shown in  Algorithm \ref{algorithm_hierarchical}. 

At every iteration, several hyperspheres are applied on the data space. In every hypersphere, the feature space is either be untouched, which results in the original feature space of $1\times d$, reduced to a subset by random sampling, which results in a random subset of feature space $1\times p$ where $p<d$, or reduced to a subset using feature selection, which results in a subset of feature space $1\times p$, where $p<d$ and features are sorted in descending order based on their correlation with labels (see line 11 in Algorithm \ref{algorithm_hierarchical}). 

 We refer to the number of instances as the data space and to the dimensions of the data as the feature space. As the algorithm suggests, instances can be sampled from within hyperspheres using stratified sampling (see line 12 in Algorithm \ref{algorithm_hierarchical})\cite{barnett1974elements}. Collecting all the sampled points and selected features from hyperspheres, we then apply a feature extraction method explained in section \ref{section_literature_review}. The weights resulting from the feature extraction method of our choice are stored for transforming the test dataset later, and are used to project the whole data into the concluded subspace at the end of iteration (see lines 15 and 16 in Algorithm \ref{algorithm_hierarchical}).

Note that in the Fisher Discriminant Analysis (FDA) method, we slightly strengthen the diagonal of the coefficients matrix denoted by $W$ in order to avoid data collapsing into subspace due to low ranks. Bear in mind that it does not have any effect on the projection directions that we desire. 

At every iteration, the number of hyperspheres, denoted by $n_s$, decreases with respect to iteration index while the radius of the hyperspheres, denoted by $r$, increases. This is because as the algorithm progresses, we want to explore more of the data space and see all data points at least once, without having much overlap among the sampling area. If feature selection is also applied, the subset portion of features denoted by $n_f$ is altered with respect to the iteration index because as we move into a new space, new features are being created during feature extraction. The instance's stratified sampling portion also decreases at each iteration because as the radius of hyperspheres increases, more data can fit in while we only want to sample a portion of each available class. 

We coded the algorithm in MATLAB. The initial value of radius, number of hyperspheres, instance sampling portion, feature selection subset portion, and feature extraction subset portion are $r = 0.1 \sigma$, $n_s = \lfloor 0.01 \times n \rfloor$ (clipped to $10 \leq n_s \leq 20$), and $p_\tau = 1$, $n_f = 90\%$, and $FE = 90\%$. These values will be updated by $r = r + \Delta r$, $n_s = \max(n_s - \lceil 0.2 \times n_s \rceil, 1)$, $p_\tau = \max(p_\tau - 0.05, 0.2)$, where $\Delta r = 0.3 \sigma$ and $\sigma$ is the average standard deviation along features, $n_s = \max(n_f - \lceil 0.95 \times n_f \rceil, 50)$, and $FE = \max(FE - \lceil 0.95 \times FE \rceil, 50)$.

\SetAlCapSkip{0.5em}
\IncMargin{0.8em}
\begin{algorithm2e}[!t]
\DontPrintSemicolon
    \textbf{Procedure: } Hierarchical Subspace Learning($\b{X}$)\;
    \textbf{Input: } $\b{X}$: dataset\;
    Initialize $r$, $n_s$, and $p_\tau$\;
    \For{$\tau$ from $1$ to $T$}{
        $r := \text{increasing function of } \tau$ \;
        $n_s := \text{decreasing function of } \tau$\;
        $p_\tau := \text{decreasing function of } \tau$\;
        $n_f := \text{decreasing function of } \tau$\;
        \For{$s$ from $1$ to $n_s$}{
            $\b{c}_s \sim \text{range}(\b{X})$\;
            Feature selection : Correlation based feature selection with portion size $n_f$ within the $s$-th hypersphere\;
            Instance Sampling: Draw a stratified sample with sampling portion $p_\tau$ within the $s$-th hypersphere\;
            }
            \label{alg_sampling}
            Apply desired subspace learning method\; \label{alg_optimization}
            Store the weight matrix\;
            Project $\b{X}$ onto $\mathbb{C}\text{ol} (\b{W})$: $\b{X} \gets \b{X} \b{W}$\; \label{alg_projection}
        
    }
    
\caption{Hierarchical Subspace Learning}\label{algorithm_hierarchical}
\end{algorithm2e}
\DecMargin{0.8em}

%%%%%%%%%%%%%%%%%%%%%%%%%%%%

\begin{table*}[!ht]
% \begin{minipage}{\textwidth}
\caption{Comparing accuracies of the proposed manifold learning methods \\
in raw, non-hierarchical  and hierarchical subspace learning for classification.}
\label{table_accuracy}
\centering
% \scalebox{1}{    %%% --> for resizing tables
\begin{center}
\hspace*{-0.5cm}

\begin{tabular}{l || l || l || p{6mm} | p{6mm} | p{6mm} | p{6mm} || p{6mm} | p{6mm} | p{6mm} | p{6mm} || p{6mm} | p{6mm} | p{6mm} | p{6mm}}
&&& \multicolumn{12}{c}{Classifiers}\\
\cline{4-15}
Datasets & & Feature Extraction & \multicolumn{4}{c||}{No Feature Selection} & \multicolumn{4}{c||}{Random Feature Selection} & \multicolumn{4}{c}{Feature Selection}\\
\cline{4-15}
 & & & LDA & KNN & RF & QDA & LDA & KNN & RF & QDA & LDA & KNN & RF & QDA\\
\hline
\hline

\multirow{9}{*}{Iris} & \multirow{1}{*}{Raw} &  & 81.80 & 72.70 & 81.80 & 76.30 & 81.80 & 72.70 & 81.80 & 76.30 & 81.80 & 72.70 & 81.80 & 76.30 \\
\cline{2-15}
& \multirow{4}{*}{Original} & LDA & 81.81 & 72.70 & 81.80 & 77.70 & 81.81 & 72.70 & 81.80 & 77.70 & 81.81 & 72.70 & 81.80 & 77.70 \\
& & PCA & 72.27 & 71.00 & 81.81 & 72.72 & 77.27 & 72.71 & 81.81 & 77.80 & 77.27 & 72.70 & 82.30 & 76.20 \\
& & GDA & 81.80 & 76.00 & 86.36 & 77.65 & 81.80 & 80.80 & 86.30 & 86.30 & 81.80 & 78.18 & 91.90 & 86.30 \\
& & RICA & 81.81 & 69.09 & 65.45 & 76.30 & 81.81 & 69.09 & 65.45 & 68.81 & 81.81 & 69.09 & 65.45 & 76.30 \\
\cline{2-15}
& \multirow{4}{*}{Hierarchical} & LDA & 81.81 & 77.27 & 81.80 & 77.00 & 81.81 & 77.27 & 86.36 & 86.36 & 81.81 & 77.27 & 86.36 & 86.36 \\
& & PCA & 81.81 & 86.18 & 82.34 & 77.00 & 81.81 & 72.45 & 82.34 & 77.00 & 81.81 & 81.18 & 82.34 & 77.27\\
& & GDA & 81.80 & 87.27 & 86.36 & 77.65 & 81.80 & 80.80 & 86.30 & 86.30 & 81.80 & 83.50 & 91.90 & 86.30 \\
& & RICA & 81.81 & 72.72 & 65.45 & 76.30 & 81.81 & 72.72 & 65.45 & 61.81 & 81.81 & 72.72 & 65.45 & 76.30 \\
\hline
\hline

\multirow{9}{*}{Breast Cancer} & \multirow{1}{*}{Raw} &  & 97.14 & 97.14 & 98.09 & 99.04 & 97.14 & 97.14 & 98.09 & 99.04 & 97.14 & 97.14 & 98.09 & 99.04 \\
\cline{2-15}
& \multirow{4}{*}{Original} & LDA & 97.14 & 96.19 & 96.19 & 96.19 & 94.28 & 93.33 & 94.28 & 98.95 & 96.19 & 96.19 & 94.28 & 94.28 \\
& & PCA & 95.23 & 97.14 & 94.28 & 95.23 & 96.19 & 93.33 & 92.38 & 93.33 & 95.23 & 96.19 & 96.19 & 98.09 \\
& & GDA & 96.19 & 96.19 & 96.19 & 97.14 & 97.14 & 96.19 & 96.19 & 100 & 93.33 & 94.28 & 92.38 & 94.28 \\
& & RICA & 96.19 & 95.23 & 94.28 & 95.23 & 98.09 & 98.09 & 99.04 & 99.04 & 94.28 & 93.33 & 94.28 & 96.19 \\
\cline{2-15}
& \multirow{4}{*}{Hierarchical} & LDA & 97.14 & 98.09 & 97.14 & 97.14 & 94.28 & 95.23 & 96.19 & 93.33 & 96.19 & 96.19 & 97.14 & 97.14 \\
& & PCA & 95.23 & 98.09 & 96.19 & 95.28 & 95.23 & 96.19 & 93.33 & 94.28 & 98.09 & 97.14 & 99.04 & 96.19 \\
& & GDA & 96.19 & 98.09 & 96.19 & 97.14 & 99.04 & 98.09 & 97.14 & 98.09 & 93.33 & 94.28 & 92.38 & 94.28 \\
& & RICA & 96.19 & 96.19 & 96.19 & 95.23 & 98.09 & 98.09 & 99.04 & 99.04 & 94.28 & 96.19 & 96.19 & 96.19 \\
\hline
\hline

\multirow{9}{*}{Isolet} & \multirow{1}{*}{Raw} &  & 99.87 & 92.05 & 93.58 & 95.71 & 99.87 & 92.05 & 93.58 & 95.71 & 99.87 & 92.05 & 93.58 & 95.71 \\
\cline{2-15}
& \multirow{4}{*}{Original} & LDA & 84.35 & 82.30 & 82.05 & 81.53 & 81.70 & 81.53 & 78.97 & 79.23 & 83.58 & 82.82 & 78.46 & 81.02 \\
& & PCA & 91.02 & 83.83 & 75.20 & 81.02 & 81.02 & 74.10 & 81.02 & 81.02 & 81.53 & 81.79 & 79.74 & 91.79 \\
& & GDA & 79.23 & 77.14 & 77.14 & 85.89 & 76.41 & 73.84 & 71.79 & 73.58 & 76.12 & 72.84 & 71.02 & 73.30 \\
& & RICA & 77.94 & 71.28 & 84.35 & 80.00 & 71.53 & 73.07 & 77.69 & 81.53 & 93.33 & 74.35 & 83.35 & 72.30\\
\cline{2-15}
& \multirow{4}{*}{Hierarchical} & LDA & 91.28 & 88.46 & 88.46 & 81.53 & 80.00 & 74.35 & 76.92 & 75.89 & 89.48 & 88.17 & 89.48 & 88.17 \\
& & PCA & 93.07 & 83.05 & 89.74 & 94.87 & 78.00 & 70.60 & 75.21 & 75.00 & 86.64 & 88.36 & 83.33 & 84.61 \\
& & GDA & 85.41 & 86.41 & 82.30 & 92.10 & 63.07 & 67.17 & 57.69 & 63.39 & 81.53 & 80.51 & 80.25 & 84.10 \\
& & RICA & 77.43 & 71.28 & 79.74 & 80.51 & 80.00 & 77.69 & 79.48 & 80.00 & 93.33 & 79.23 & 78.97 & 73.58 \\
\hline
\hline

\multirow{9}{*}{MNIST} & \multirow{1}{*}{Raw} &  & 75.00 & 80.00 & 88.00 & 61.00 & 75.00 & 80.00 & 88.00 & 61.00 & 75.00 & 80.00 & 88.00 & 61.00 \\

\cline{2-15}
& \multirow{4}{*}{Original} & LDA & 71.00 & 72.00 & 62.00 & 69.00 & 74.00 & 72.00 & 63.00 & 69.00 & 71.00 & 73.00 & 65.00 & 67.00 \\
& & PCA & 75.00 & 81.00 & 84.00 & 68.00 & 75.00 & 80.00 & 82.00 & 68.00 & 75.00 & 81.00 & 84.00 & 68.00 \\
& & GDA & 70.00 & 79.90 & 65.00 & 58.90 & 70.00 & 79.90 & 65.00 & 58.90 & 70.00 & 79.90 & 65.00 & 58.90 \\
& & RICA & 71.00 & 70.00 & 73.00 & 64.00 & 71.00 & 67.00 & 68.00 & 57.00 & 71.00 & 62.00 & 66.00 & 64.00  \\
\cline{2-15}
& \multirow{4}{*}{Hierarchical} & LDA & 74.00 & 74.00 & 70.00 & 69.00 & 74.00 & 72.50 & 65.00 & 68.00 & 73.00 & 74.00 & 69.00 & 73.00 \\
& & PCA & 80.00 & 87.00 & 82.00 & 67.00 & 79.00 & 80.00 & 84.00 & 69.00 & 80.00 & 84.00 & 86.00 & 70.00 \\
& & GDA & 74.00 & 79.00 & 68.00 & 60.00 & 58.00 & 45.00 & 58.00 & 47.00 & 80.00 & 79.00 & 70.00 & 76.00 \\
& & RICA & 71.00 & 74.00 & 75.00 & 75.00 & 68.00 & 60.00 & 62.00 & 72.00 & 71.00 & 74.00 & 64.00 & 64.00 \\
\hline
\hline

\multirow{9}{*}{ORL Faces} & \multirow{1}{*}{Raw} &  & 92.50 & 87.50 & 92.50 & 89.90 & 92.50 & 87.50 & 92.50 & 89.90 & 92.50 & 87.50 & 92.50 & 89.90 \\
\cline{2-15}
& \multirow{4}{*}{Original} & LDA & 85.00 & 88.75 & 67.50 & 82.20 & 85.00 & 88.75 & 70.00 & 75.00 & 85.00 & 88.70 & 71.20 & 80.20 \\
& & PCA & 92.50 & 83.75 & 83.75 & 87.50 & 92.50 & 83.40 & 83.75 & 86.00 & 92.50 & 84.00 & 81.25 & 88.67 \\
& & GDA & 74.00 & 72.13 & 63.00 & 52.22 & 73.50 & 66.32 & 63.00 & 52.22 & 73.70 & 68.23 & 63.00 & 52.22 \\
& & RICA & 73.75 & 75.00 & 83.75 & 81.25 & 73.75 & 75.00 & 82.50 & 81.25 & 73.00 & 75.00 & 85.00 & 81.25 \\
\cline{2-15}
& \multirow{4}{*}{Hierarchical} & LDA & 85.60 & 88.75 & 70.86 & 84.00 & 83.50 & 85.00 & 65.30 & 69.00 & 88.00 & 89.70 & 81.45 & 82.43  \\
& & PCA & 92.50 & 84.50 & 84.50 & 90.12 & 58.50 & 76.45 & 73.50 & 67.90 & 82.50 & 85.30 & 85.42 & 89.00 \\
& & GDA & 78.54 & 72.13 & 66.60 & 60.00 & 67.50 & 66.25 & 56.73 & 57.00 & 80.00 & 76.25 & 67.50 & 60.50 \\
& & RICA & 73.75 & 75.00 & 83.75 & 81.25 & 70.86 & 70.86 & 80.00 & 79.40 & 76.07 & 79.92 & 85.00 & 83.12 \\
\hline
\hline

\end{tabular}%
\end{center}
% }
% \end{minipage}
\end{table*}

\section{Experimental Results and Analysis}\label{section_experiments}

\subsection{Datasets and Setup}
We used five publicly available datasets in this paper with their specifications presented in Table \ref{dataset_specification}. The first dataset is Fisher Iris \cite{web_UCI_repository}, with 150 samples in three classes with four features. The second dataset used is ORL faces \cite{web_ORL_dataset}, which includes a set of images of 40 distinct individuals, each having 10 images with the size of $112\times92$ pixels. Third, we used the Isolet dataset \cite{Dua:2019}, which is a voice dataset for alphabet classification. It includes 150 participants repeating each letter twice to make a total of $7,797$ instances with $617$ attributes. A subset of MNIST \cite{lecun1998gradient} was used with $3,000$ images, each $28\times28$ pixels of handwritten digits between zero and nine. Another dataset used in this paper is the Wisconsin breast cancer diagnostic dataset \cite{Dua:2019}, with $569$ instances and $32$ features. This paper tries to test a variety type of datasets with a range of features to show the credibility of the proposed method. 

All datasets are divided to 75\%-15\%-15\% train, validation and test sub-data. The ORL dataset was further projected to $38$ leading eigenfaces.

\begin{table*}[!h]
% \begin{minipage}{\textwidth}
\caption{Datasets specifications}
\label{dataset_specification}
\centering
% \scalebox{1}{    %%% --> for resizing tables

\begin{tabular}{ || l || l || l || l || l || }
\hline
Datasets & Size & No Classes & Type & Task\\
\hline
\hline
Iris & $150 \times4$& 3 & Digit & Pattern Recognition \\
Breast Cancer & $569 \times 32$ & 2 & Digitized Image & Diognostic \\
Isolet & $7797 \times 617$ & 26 & Digitized Voice & Voice Recognition\\
MNIST & $3000 \times 784$ & 10 & Image, Pixel & Digit recognition\\
ORL & $400 \times 1178$ & 10 & Image, Pixel & Face Recognition\\
\hline
\hline

\end{tabular}%
%  }
% \end{minipage}
\end{table*}

\subsection{Comparison of Manifold Learning Methods in the Non-Hierarchical and Hierarchical Approaches}

Each dataset is tested in three categories: no feature selection, random feature selection, and chi-squared feature selection. For each of these methods, the algorithm goes through manifold learning methods including PCA, LDA, ICA and GDA. Four different classifiers are used to compare the accuracy of these methods. Eventually, these results are compared to raw data going through a classifier without any dimensionality reduction and the original feature extraction method applied once. Note that Euclidean distance is the metric used for the kNN classifier. Table \ref{table_accuracy} represents all the classification accuracies for Iris, Breast Cancer, Isolet, MNIST and ORL faces datasets. 

In almost all datasets, the hierarchical approach results in higher accuracy. Exceptions occur in the Iris and the Breast Cancer datasets with many compatible accuracies due to the simplicity of the datasets and their low number of features. These cases can easily be handled in a non-hierarchical approach. As the number of attributes increases, the increase between original method accuracy to the hierarchical approach is clearer. The hierarchical approach accuracy exceeds the raw data accuracy in some cases such as Breast cancer - LDA  for kNN classifier, or MNIST - PCA for the LDA classifier. This is due to the curse of lower dimensionality in which higher dimensions do not always mean higher accuracy. This can happen in cases where the number of features sampled is greater than the number of instances sampled. 

Random subset selection, however, does not guarantee better results since it considers neither the between-feature correlation nor the feature-label correlation. But, as can be seen in Table \ref{table_accuracy}, the iterative nature of the method makes up for this random feature selection resulting in comparable accuracies with the original method. Omitting feature selection and having only feature extraction as the dimensionality reduction, results in higher accuracies that outperform both random feature selection and correlation based feature selection, as expected. This is because while the goal of dimensionality reduction is achieved by the feature extraction, we still keep more of our feature space by not utilizing feature selection. This approach, however, takes  more run-time than others because the projection direction matrix, which depends on the number of features, is bigger. 

Looking closely at Table \ref{table_accuracy}, it can be seen that LDA feature extraction outperforms other dimensionality reduction approaches, but it is closer to PCA than the rest. This is due to LDA being a supervised method using the training labels while PCA is unsupervised and does not require any labels. The comparability is because the datasets are more linear. PCA and LDA are both linear methods that match well with linear datasets. GDA is an extension of LDA and hence is expected to have comparable results. 

Moreover, the Random Forest classifier achieves higher accuracies on average because of creating a forest of trees and taking a majority vote. In this way, the probability of error is minimized. Besides the RF classifier, the LDA classifier also presents high accuracies, especially in cases of LDA feature extraction. This is again because the data is linear and hence LDA classifier pairs well with LDA dimensionality reduction. 

The main tradeoff in the hierarchical approach is between accuracy and run time. The run time in smaller datasets such as Iris and Wisconsin Breast Cancer is not significantly different. However, as the dimensionality of the dataset increases, especially with a greater number of features, the run time increases to an average of 297.12, 154.56, and 145.31 seconds for the Isolet, MNIST and ORL faces datasets, versus 46.42, 20.60, 20.22 seconds respectively for the baseline. This increase is predictable due to the hierarchical nature of the algorithm and the dependence of a classifier’s time and space complexity on the input matrix at every iteration.
\newline

\section{Conclusion and Future Direction}\label{section_conclusion}

Dimensionality reduction is often used to create lower-dimensional spaces that represent data in a more visual manner to make exploration easier. It is commonly used as a pre-processing method in machine learning and statistics. However, these methods do not always guarantee higher classification accuracies. Inspired by the hierarchical approach for large margin metric learning using stratified sampling \cite{poorheravi2020acceleration}, we introduce a hierarchically iterative approach as a pre-processing method to increase the accuracy of classification. Five publicly available datasets are tested through three different types of the algorithms: without feature selection, with random feature selection and with correlation-based feature selection. For each method, four different feature extraction methods are used. Finally, three types of results are gathered: classifying raw data, classifying the original feature extracting method, and the classifying data pre-processed using the proposed hierarchical approach.

An accuracy comparison summary table between the original approach and the proposed hierarchical approach is presented in Table \ref{summary_table} using $No$ $Feature$ $Selection$ and $Feature$ $Selection$ methods. The table shows up to an average of 3\% increase in accuracy for the hierarchical approach at the expense of run time. An approach for possible future work can be testing non-linear manifold learning methods such as kernel PCA or kernel LDA using this method since such methods initially expand the feature space using a kernel to achieve a non-linear mapping to maximize the variance while applying dimensionality reduction. 
% \newline

\begin{table*}
\caption{Summary of Table \ref{table_accuracy} Average Accuracy}
\label{summary_table}
\centering
\begin{tabular}{|| l || l || l || l || l || l ||}
\hline
\multirow{2}{*}{Feature Extraction} & & \multicolumn{4}{c ||}{Classifiers}\\
\cline{3-6}
 & & LDA & KNN & RF & QDA\\
% \cline{4-15}
\hline
\hline
\multirow{2}{*}{LDA}&\multirow{1}{*}{Original} & 83.68 &82.53 & 78.02 & 80.68\\
& \multirow{1}{*}{Hierarchical} & 85.83 & 85.19 & 83.16 & 83.57\\
\hline
\multirow{2}{*}{PCA} &\multirow{1}{*}{Original} & 84.75 & 83.24 & 84.25 & 82.72\\
& \multirow{1}{*}{Hierarchical} & 87.16 & 87.48 & 87.09 & 84.13 \\
\hline
\multirow{2}{*}{GDA} &\multirow{1}{*}{Original} & 79.61 & 79.29 & 77.09 & 73.68\\
& \multirow{1}{*}{Hierarchical} & 83.26 & 83.64 & 80.14 & 78.80 \\
\hline
\multirow{2}{*}{RICA} &\multirow{1}{*}{Original} & 81.41 & 75.43 & 79.49 & 78.68\\
& \multirow{1}{*}{Hierarchical} & 81.66 & 79.12 & 78.97 & 80.14 \\

\hline
\hline

\end{tabular}%
%  }
% \end{minipage}
\end{table*}

\bibliographystyle{IEEEtran}
\bibliography{References}

% that's all folks
\end{document}